\documentclass{article}
\usepackage[margin=1in]{geometry}
\usepackage{inputenc}
\usepackage[disable]{todonotes}
\usepackage{natbib}
\usepackage{graphicx}
\usepackage{authblk}
\usepackage{amsmath}
\usepackage{amssymb}
\usepackage{float}
\usepackage{subcaption}
\usepackage{adjustbox}
\usepackage{hyperref}
\title{Adversarial Training with Ladder Networks}
\author[2]{Juan Maro\~nas Molano \\ jmaronasm@gmail.com}
\author[1]{Alberto Albiol Colomer \\ alalbiol@iteam.upv.es} 
\author[2]{Roberto Paredes Palacios \\  rparedes@dsic.upv.es }
\affil[1]{Departamento de Comunicaciones, Universitat Polit\`ecnica de Val\`encia}
\affil[2]{Departamento de Sistemas Inform\'aticos y Computaci\'on, Universitat Polit\`ecnica de Val\`encia}

\usepackage{datetime}
\newdate{date}{01}{03}{2017}
\date{\displaydate{date}}

\begin{document}

\maketitle
\begin{abstract}
    
The use of unsupervised data in addition to supervised data has lead to a significant improvement when  training discriminative neural networks. However, the best results were achieved with a training process that is divided in two parts: first an unsupervised pre-training step is done for initializing the weights of the network and after these weights are refined with the use of supervised data. Recently, a new  neural network topology called Ladder Network, where the key idea is based in some properties of hierarchichal latent variable models, has been proposed as a technique to train a neural network using supervised and unsupervised data at the same time with what is called semi-supervised learning. This technique has reached state of the art classification. On the other hand adversarial noise has improved the results of classical supervised learning. In this work we add adversarial noise to the ladder network and get state of the art classification, with several important conclusions on how adversarial noise can help in addition with new possible lines of investigation. We also propose an alternative to add adversarial noise to unsupervised data.

\end{abstract}

\begin{section}{Introduction}

Learning unconditional distributions $p(x)$ using unsupervised data can be useful for learning a neural network that models a conditional distribution $p(t|x)$ where $t$ represents the target of the task. Learning such distributions can be used to initialize the weights of the networks, using the supervised data to refine the parameters and adjust them to the task at hand. This pre-trained neural network has already learnt important features to represent the underlying distribution of the data $x$. \\

Classical approaches for learning $p(x)$ and use them for then learning $p(t|x)$ are under two subsets. Deep belief networks \citep{hinton2006reducing} are based on training pairs of Restricted Boltzmann Machines (RBM), which are a kind of probabilistic energy-based model (EBM) \citep{lecun-06}\citep{Lecun05lossfunctions}, and then perform a finne-tunning of the parameters. Deep boltzmann machines \citep{salakhutdinov2009deep} are EBM with more than one hidden layer to create a deep topology to after perform a fine tunning. On the other hand autoencoders \citep{Hinton:1990:CLP:120048.120068} \todo{Revisarme esta referencia porque de autoencoders encuentro mucho} are neural networks where the output target is the input. The autoencoder is divided in two parts: encoder and decoder. The encoder takes the input $x$ and start reducing the dimensionality where each hidden layer $h$ of the neural network represents a dimension. The decoder has the same topology of the encoder but starts from the last layer of the encoder (is shared between both parts) and perform operations to have an output $\bar{x}$ which should be as closed as possible to the input. The autoencoder is trained to achieve this property by minimizing the sum of squared error between the input and the reconstruction, as it is assumed the error distribution between input and prediction is gaussian. We then take the pretrained encoder and perform a finne-tunning of the parameters using supervised data.\\

\end{section}

\begin{section}{Semi Supervised Learning and Ladder Networks}
Semi supervised learning implies learning $p(x)$ and $p(t|x)$ at the same time. This means using supervised and unsupervised data in the same learning procedure. The key idea of semi supervised learning is that unsupervised learning should find new features that correlates well with the already found features suitable for the task. This suitability is driven by the supervised learning procedure.\\ %\todo{Aqui no se si citar a rasmus y valpola. La verdad es que lo lei ahi pero tampoco parece un descubrimiento suyo. Además yo no se hasta que punto se puede demostrar que al mezclar ambos procesos a la vez lo que hace el aprendizaje no supervisado es buscar caracteristicas que correlen con las que el supervisado encuentra ya que el objetivo del no supervisado es directamente p(x)}\\

Mixing this learning schemes can end up stalling the learning procedure for the fact that the targets of the learning schemes are different. On one side unsupervised learning tries to encode all the necessary information for reconstructing the input. On the other supervised learning is more focused on finding abstract and invariant features (at different levels of invariability) for discriminate the different inputs. Unsupervised features such as relative position or size in a face description maybe not necessary for a discriminative task and discriminative features tipicaly do not have information about data structure so are unseful to represent $x$. \\

To perform semi supervised learning the key idea is that unsupervised learning should be able of discard information necessary for the reconstruction, and encode this information in other layer. For example suppose supervised learning finds useful to have a characteristic in a layer $\underset{l}{h}$ where $l$ represent a particular layer. At this level unsupervised learning needs some kind of representation to keep the reconstruction error low and this information is unuseful for supervised learning performance. If unsupervised learning could be able of representing this information in another level $l'$ we could still keep the reconstruction error low and supervised learning could have the information needed at this level.\\

\begin{subsection}{Latent Variable Models}
Latent variable models are models for learning unconditional probability distributions with the particularity that given a latent variable we can reconstruct the observed variable by means of a likelihood probability distribution. This means the proceedure not only depends on $h$ but also on a random procedure so in someway the model adds their own bits of information for the reconstruction. More formally for a discret distribution:

\begin{equation}
p(x)=\underset{\forall h}{\sum}p(x|h)\cdot p(h)
\end{equation}

where $p(x|h)$ can be modelled like:

\begin{equation}
x=f_\theta(h) + n_{\theta_n}
\end{equation}

that is the mean of the likelihood distribution is given by the projection of $h$ to the observed space and the deviation is given by some noise process. We can learn the parameters using EM \citep{Dempster77maximumlikelihood}. The main bottleneck of EM in some latent variable models (like RBM) is that implies computing the posterior probability $p(h|x)$ of the latent variable, which is sometimes mathematically intractable. Alternatives such as stochastic gradient guided methods are used but sometimes implies making approximations using slow methods like MCMC exploration.\\

The structure of this models fits well with the semi supervised learning paradigm because it allows discarding information for the fact that this information can be somehow added to the reconstruction. However a one hidden layer latent variable model is unable of discard information because it needs to represent all the necessary information to represent $p(x)$ in the hidden layer. The solution is to use hierarchical latent variable models where the information can be somehow distributed between the different variables and each variable is able of adding their own bits of information.\\

The two key ideas of this models are that modelling the observed variable as a probability distribution implies that a hidden variable can somehow add information (so we can discard information that is then added) and that making a hierarchy allows information discarding.

\begin{subsection}{Ladder Networks}

Ladder Network \citep{valpola2015neural} is neural network topology that implements the key idea of latent variable models that make suitable mixing supervised and unsupervised learning at the same time. The topology of the network is given by figure \ref{fig:ladderNetwork}:

\begin{figure}[H]
    \centering
    \includegraphics[scale=0.2]{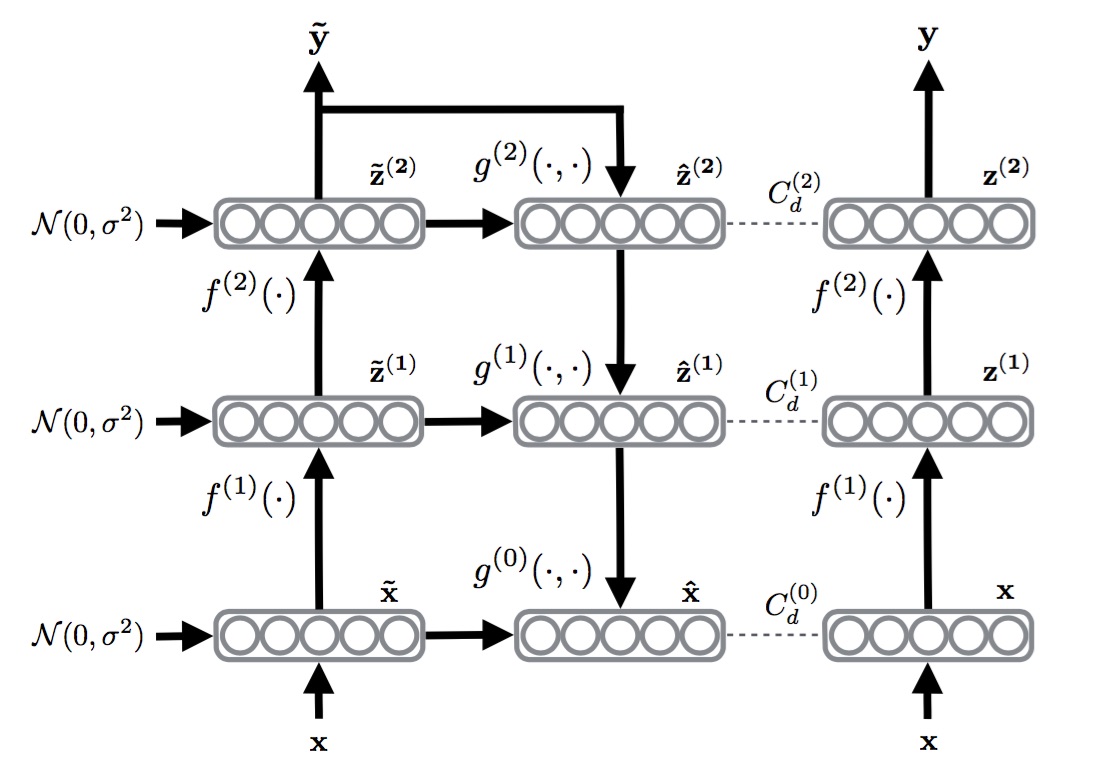}
    \caption{Ladder Network Topology \citep{DBLP:journals/corr/RasmusVHBR15}}
    \label{fig:ladderNetwork}
\end{figure}

where as we can see is a kind of autoencoder with lateral connections between the encoder and the decoder, at each level of the network. The decoder reconstruct a variable in level $\underset{l}{\bar{h}}$ using the above reconstructed variable  $\underset{l+1}{\bar{h}}$ and the corrupted variable at the same level $\underset{l}{\widetilde{h}}$ using a denoising function $g(\cdot,\cdot)$. This topology allows information discarding at any level, as long as it is needed. The reconstruction not only depends on the level above (and so depends on all the preceding levels) but also on the encoder part. This means their can be extra added information to the reconstruction (as the latent variable models did) and we can loose information at any level that is encoded in other levels. Note that with this topology any lower level can be influence by any signal in any higher level no matter if it is from the encoder or decoder path. This means that the features the neural network learns are somehow distributed through the network as long as the supervised learning refines which kind of feature need at any level. \\

The minimized cost is given by the expression:
\begin{equation}
C=C_s+C_u
\end{equation}

where subindex represent supervised and unsupervised. The supervised cost is the cross entropy, that is (for only one sample $X$):

\begin{equation}
C_s=-\frac{1}{K}\underset{k=1}{\overset{K}{\sum}}\log{P(\widetilde{Y}_k=\widehat{Y}_k|\widetilde{X})}
\end{equation}

where $K$ is the layer dimension and $\widehat{Y}$ is the target. We use the corrupted output as the target to regularize. The unsupervised cost is the weighted sum of a reconstruction measure at each level:

\begin{equation}
C_u=\omega_0\cdot(\lvert\lvert x-\bar{x}\rvert\rvert_2)^2+\underset{l=1}{\overset{L}{\sum}}\omega_d\cdot(\lvert\lvert \underset{l}{z}-\underset{l}{\bar{z}}\rvert\rvert_2)^2
\end{equation}

This unsupervised cost allows deep architectures because each parameter of the topology can be well trained and is difficult to find the gradient vanishing problem.

\begin{subsubsection}{Denoising Principle}

We should pay special attention to the added noise in the encoder. This noise serves for two purposes. The first one is implementing the denoising principle of the denoising autoencoder \citep{Vincent:2008:ECR:1390156.1390294}\citep{Vincent:2010:SDA:1756006.1953039} which serves as a good regularizator. \\

Noise is also added to force the $g(\cdot,\cdot)$ function to use the information in the above layer to reconstruct the signal because the signal which minimizes the cost at a level $l$ is just $\underset{l}{h}$. This avoid the reconstruction function just copy the encoded signal to the decoder. One more special thing to remark is that the ladder network uses batch normalization for two purposes. First is avoid internal covariate shift \citep{ioffe2015batch} and the other is because we have to avoid the encoder just output constant values, as these are the easiest ones to denoise \citep{DBLP:journals/corr/RasmusVHBR15}.

\end{subsubsection}

\end{subsection}
\end{subsection}
\end{section}

\begin{section}{Adversarial Noise}

Suppose we take a training data $X$ from our distribution. $X$ is an 8-bit normalized image which means there is a precision error of $\frac{1}{255}$. Now we convert this image to floating point and add a perturbation $\tau<\frac{1}{2\cdot 255}$. If we then convert this new corrupted sample $\widetilde{X}=X+\tau$ to an 8-bit image it is clear that $X=\widetilde{X}$ because the perturbation lies in the same quantification interval. Our model should correctly classify this sample. \citep{szegedy2014intriguing} discover that neural networks are not robust to adversarial examples, that is, examples nearly similar but that highly increase the misclassification error.\\

The first thing we did is see if the ladder network was robust to adversarial examples. For that reason we trained one model which had a 0.51\% error on the test set. We then corrupt the test set with adversarial noise and with random noise ensuring the power from both noises was the same. \\

Adversarial examples where computed, following \citep{goodfellow2014explaining}, like:

\begin{equation}
\widetilde{X}_a=X+\tau\cdot\text{sign}\{\frac{\partial C_s}{\partial x}\big|_X\}
\label{equ:adv_sign}
\end{equation}

where $\tau$ ensures the perturbation $\epsilon$ is under the quantification error. This fast computing methods can be derived from a first order Taylor series approximation of $C_s(x+\epsilon)$, see chapter 1 from \citep{advperturbGood}. Note that the energy of the adversarial noise is $\tau\cdot K$ where $K$ is the input dimension and we can corrupt a sample using random gaussian noise like:

\begin{equation}
\widetilde{X}_r=X+\tau\cdot\text{sign}\{q\}
\end{equation}

where $q$ is a samples of dimensionality $K$ drawn from a zero mean gaussian distribution $\mathcal{N}(0,I)$. The power of these noises is the same and is $\tau\cdot K$. The next figure shows a sample of the MNIST test set corrupted with both types of noise. As we see we cannot say which sample will suppose a higher missclassification error. But we can see the differences in the test error. We conclude the ladder network is not robust to adversarial examples.

\begin{figure}[H]
    \begin{subfigure}{.5\textwidth}
        \centering
        \includegraphics[width = 0.5\textwidth]{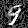}
        \caption{$X_a$. Error percentage 6.87\%}
    \end{subfigure}%
    \begin{subfigure}{.5\textwidth}
        \centering
        \includegraphics[width = 0.5\textwidth]{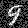}
        \caption{$X_r$. Error percentage 0.61\%}
    \end{subfigure}
\end{figure}

\begin{subsection}{Computation of Adversarial Noise}

Adversarial examples can be generated in several ways. An expensive way to generate adversarial examples is by box-constrained L-BFGS \citep{szegedy2014intriguing}. A fast way, call the fast gradient method \citep{goodfellow2014explaining}, compute adversarial examples adding the sign of the derivative of the cost function respect to the input, following equation \ref{equ:adv_sign}.\\

This correspond to the solution of a maximization problem whose details can be found in \citep{advperturbGood}. The solution of the problem ensures $||\epsilon||_{\infty}<\tau$. In this work we explore another way of doing this. We generate adversarial noise following the next equation (we will refer to norm adversarial):

\begin{equation}
\widetilde{X}=X+\tau_a\cdot\frac{\partial C_s}{\partial x}\big|_X;\big\lvert\big\lvert \frac{\partial C_s}{\partial x}\big|_X \big\rvert\big\rvert_2=1
\end{equation}

Note that setting the norm of the gradient vector to be one we are also ensuring $||\epsilon||_{\infty}<\tau$ because the norm of this vector is upper-bounded by the norm of the sign gradient. The norm of the sign gradient is $K^{\frac{1}{2}}$ where $K$ represent the dimension of that particular layer. By setting different $\tau$ values for the two different noise computation we can have adversarial noise with similar powers. For the MNIST case the sign adversarial noise would have a $784$ norm value and that means the norm adversarial power is one magnitud order below the sign adversarial. \\

We will see that this way of computing adversarial noise makes sense for three reasons. First one is because in the hyperparameter that sets the power of adversarial noise we first look for parameters changing only the magnitud order.  The second one relies on the intuitive idea behind adding the gradient of the cost function respect to the input. Note that if we add the gradient, which is by definition  the direction where the cost maximally changes respect to an input to that cost, we are adding a perturbation which directly increase the cost. However if we perform a sign operation we are afecting the phase of this vector and we are not exactly adding the adversarial noise in the gradient direction. The last one  will make sense after reading the next subsection but it is directly related to the preceding reason. Basically relies on how can we improve the performance of minimizing the MSE cost function.

\end{subsection}

\begin{subsection}{Unsupervised Adversarial Noise}

It is clear that the adversarial noise is focus on meausuring how robust a discriminative network is to little perturbations in sensible directions guided by the target of our task. This is the reason for computing adversarial examples using the derivative of the supervised cost, apart from the mathematics exposed in \citep{advperturbGood} intuitively we add a perturbation in the direction in which the cost highly increase. Adversarial noise is then applied to purely supervised learning. \\

    However we can easily add adversarial noise to unsupervised data taking the $\text{argmax}\{\cdot\}$ of the softmax output and use it as the true tag for that sample. It is clear that in the first steps of the optimization process that tag would be far from the true tag and that means the adversarial noise will not push towards the direction in which the cost is increase. This technique has already been proposed in \citep{miyato2016adversarial} but with a different objective. \citep{miyato2016adversarial} propose a new objective function based on the Kullback Leibler divergence of the clean and corrupted virtual adversarial VA distribution.\\ 

In this work we generate unsupervised adversarial, UA, noise following this approach and add it to the unsupervised samples. This means adversarial noise will in someway resemble gaussian noise in the early training stages. But as long as the network is well trained the adversarial noise will push the samples towards the decision bound. We can see this in figure \ref{fig:unsupervisedAdversarial}.

\begin{figure}[H]
    \centering
    \includegraphics[scale=0.5]{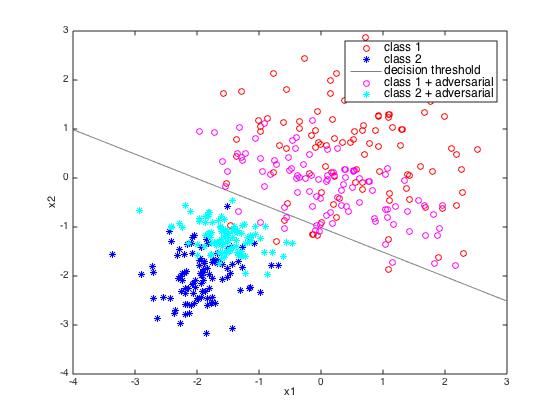}
    \caption{This figure represents two supervised data sets (dark red, circles, and dark blue,stars) drawn from two 2D gaussian distributions and the decision bound. We have applied adversarial noise to both data set computing the light red and light blue datasets. As we can see both data sets are pushed towards the decision bound and thus towards each other.}
    \label{fig:unsupervisedAdversarial}
\end{figure}

In this case we present a supervised data set to motivate our idea for unsupervised data. As we see samples correctly classified are pushed towards the decision threshold. We train adding UA noise to force the network learn to denoise samples in sensible directions. This means extracting features to represent $p(x)$ biasing in someway how we want our network to represent that distribution. The key is extract features that correlates with ones find by the supervised task reinforcing the learning in the sensible direction.\\

On the other hand, bad classified samples are also pushed towards the decision threshold, and that means they are pushed towards the data set they belong to. This is an effect possibly present when adding gaussian noise. Note that it makes more sense to add the normalized adversarial noise, as it keeps good information about the phase of the gradient vector so we are exactly pushing the samples towards the worst direction.

\end{subsection}

\begin{subsection}{Multilayer Adversarial Noise}
\label{subsec:advComputation}
One key thing has not been explored is the computation of adversarial noise at each level of the hierarchy. Understanding this concept is simple, as it only require to compute the derivative of the cost respect to the input of each layer $\underset{l}{x}$.\\

However, this highly increase the cost of each forward-backward operation in the training proceedure. To understand this, consider the input to a $L$-layer neural network. In this case we need $2\cdot L$ operations between layers to compute the gradient of the cost respect to the input and $2\cdot L$ to perform the parameter update. It is clear that we cannot compute the gradient respect to each input in the same forward-backward operation because in the final forward-backward, once we have add the adversarial noise to the previous layer, the adversarial noise of the next layer would not longer be the exact derivative respect to that input as we compute it with an input coming from a clean projection. \\

Consider the input to the network $x$ and the input to the first layer, $\underset{1}{x}=\mathcal{A}\{W\cdot x + b\}$, where $\mathcal{A}\{\cdot\}$ represents the non-linear operation such as the rectifier linear unit. The adversarial noise respect to the input could be computed as in equation \ref{equ:adv_sign} and the same operation holds for the adversarial respect to $\underset{1}{x}$. However because of the relation between $\underset{1}{x}$ and $x$ once we add some kind of perturbation to $x$, the derivative respect to $\underset{1}{x}$ changes and thus do the adversarial noise. This implies performing $2\cdot(L-l)$ between layer operations (BN,dropout,matrix product, vector addition, activation function...) to add adversarial noise to a particular layer. So for a $L$ layer network we would need $2\cdot L+\underset{l=0}{\overset{L}{\sum}}[2\cdot L-l]$ to perform one minibatch update. \\

However, we can approximate this by simply computing the adversarial noise in the first forward backward. When having saturating non-linearities like the sigmoid function, our hypothesis is that the result is the same as computing it correctly because the output of the sigmoid function is basically the same for inputs in the saturated part of the input range. Moreover, if we use dropout or gaussian noise to regularize we would have to save the dropout mask or the generated gaussian noise, add the adversarial noise, and then perform the parameter update using the same mask or number generation. This could highly increase the memory cost of a possible application. A fast approximation is computing the adversarial noise from the clean path and add it to the corrupted path, or computing it in the first feed-forward of the corrupted path and then perform another corrupted feed-forward adding the adversarial noise and generating new random numbers. 

%AQUI PONER QUE SERIA LO MISMO CON GAUSSIANO Y NO VAMO S A GUARDAR EN MEMORIA LA MASCARA DE GAUSSIANO, O DE DROPOUT. 

\end{subsection}

\end{section}

\begin{section}{Experiments}

We perform preliminar experiments over MNIST. For this dataset we evaluated a fully connected network with 50, 100, 1000 and all the labels and a convolutional network with 100 labelled data. Details on these networks topologies can be found in \citep{DBLP:journals/corr/RasmusVHBR15}.\\

\begin{subsection}{Experimental proceedure}

We have perform several experiments to validate the different proposals along the work. For the hyperparameter search of each experiment what we have done is evaluate the test set training a model with one random seed and then perform several experiments with different seeds but the hyperparameter chosen with the first seed and report the mean and standard deviation. This is not a good way to search hyperparameters but the high cost of training the ladder network made us take this solution. The code is available at \href{https://github.com/jmaronas/}{https://github.com/jmaronas/} where there is an explanation to replicate the different experiments (including the hardware and software resources). This code starts from the code of the ladder network, see \citep{DBLP:journals/corr/RasmusVHBR15} for details.\\

Some conclusions are taken from the fully connected experiments and directly applied to the convolutional models because of the high cost of the convolutional network. We validate our hypothesis by performing experiments computing the norm and sign adversarial noise. As we will see, normally, the norm adversarial noise fits better with the trained model.\\

For hyperparameter searching we first look for adversarial noise power going from $k\cdot10^0$ to $k\cdot10^{-9}\,\,k=1$. The best hyperparameter is then used to train 10 models with different seeds. If we get better results than the ladder network we stop, if not, we refine the $k$ value until we get a good result. We divide the different experiments in different sections. We will know go over our work, reporting the different results and conclusions.

\end{subsection}

\begin{subsection}{Supervised Input Adversarial}

In this first experiment we simply add adversarial noise to the input of the network. We search hyperparameters for the fully connected model under the next conditions. For each of this condition we tried both the sign and norm adversarial noises. The operations are related to the labelled data, that is, the unsupervised part of the ladder network (computation and optimization) is exactly the same. We add adversarial noise plus gaussian noise (MN-2), only adversarial noise to the input (MN-1) and gaussian to the remaining layers (as the noise in the hidden layers is used as a regularizator like dropout \citep{Srivastava:2014:DSW:2627435.2670313}) and only adversarial to labelled data (MN-1.1), that is, we take out the gaussian noise addition of the labelled data. Note that the only method that add purerly adversarial noise is the MN-1.1, as the MN-2 and MN-1 are not exactly adversarial for the reasons we exposed in section \ref{subsec:advComputation}. In general we found that MN-1.1 went better than MN-1 in all the experiments except the 50 tag, and that made us sense. However we found that this configurations were far from the state of art. We saw that the MN-2 approach yield better results than the state of art concluding that the ladder network is really sensible to gaussian noise and that we need other regularizator methods such as adding noise and that an approximate way of adding adversarial noise give also good results (we did compute the adversarial noise from the clean path). In \citep{goodfellow2014explaining} they use dropout, but do not specify if they save the dropout mask. Saving or not this mask is something to be explored. \todo{ESTO PONER EN TRABAJO FUTURO, COMO LO DEL ARTICULO DE ADVERASRIAL ANALYSISN QUE ESTOY HACIENDO}\\

The next table shows the results for the fully connected and the convoluctional networks. In the convolutional network we directly explore the MN-2 option. The table show whether we use the sign (s) or the norm (n) adversarial and the factor used:
\begin{table}[H]
\centering
\begin{adjustbox}{max width=\textwidth}
\begin{tabular}{|l|l|l|l|l|}
\hline
\multicolumn{5}{|c|}{\textbf{FC MNIST}}                                                        \\ \hline
\textbf{\# of labels} &                 50 (n;$10^{-3}$)      & 100 (n;$1.5\cdot10^{-2}$) & 1000 (n;$2\cdot10^{-7}$) & All labels (s;$3\cdot10^{-6}$)                          \\ \hline
DBM, Dropout  \citep{Srivastava:2014:DSW:2627435.2670313}  &                          &      &      & 0.79\%                                \\ \hline
Adversarial \citep{goodfellow2014explaining}                &              &      &      & 0.78\%                                 \\ \hline
Ladder full  \citep{DBLP:journals/corr/RasmusVHBR15}    &    1.750\% $\pm$ 0.582     & 1.111\% $\pm$ 0.105 & 0.880\% $\pm$ 0.075& 0.654\% $\pm$ 0.052 \\ \hline
\textbf{Adversarial Ladder full} & 1.417\% $\pm$ 0.539  & 1.110\% $\pm$ 0.055 &  0.861\% $\pm$ 0.071 & 0.639\% $\pm$ 0.056 \\ \hline
\end{tabular}
\end{adjustbox}
\caption{Results for fully connected MNIST supervised Adversarial}
\label{tab:MNISTFC}
\end{table}

\begin{table}[H]
\centering
\begin{adjustbox}{max width=\textwidth}
\begin{tabular}{|l|l|l|l|}
\hline
\multicolumn{2}{|c|}{\textbf{Convolutional MNIST}}                                           \\ \hline
\textbf{\# of labels}                 & 100 (n;$10^{-3}$)                                  \\ \hline
EmbedCNN       \citep{weston2012deep}                       & 7.75\%                                  \\ \hline
SWWAE         \citep{zhao2015stacked}                        & 9.17\%                            \\ \hline
Conv-Small, $\Gamma$-model \citep{DBLP:journals/corr/RasmusVHBR15}          & 0.938\%  $\pm$ 0.423                                \\ \hline

\textbf{Adversarial Ladder $\Gamma$-model}          &  0.884\% $\pm$ 0.427                                         \\ \hline
\end{tabular}
\end{adjustbox}
\caption{Results for Convolutional MNIST supervised Adversarial}
\label{tab:MNISTCONV}
\end{table}

\end{subsection}

\begin{subsection}{Unsupervised Adversarial}

In this subsection we present the experiments with the unsupervised adversarial approach. These experiments are computed with the approximation in the multilayer adversarial computation commented. On one part we only perform one forward-backward and on the other we compute the adversarial from the clean path.\\

The objective of this approach is first show if the unsupervised adversarial help unsupervised task, such as minimizing the MSE for the previous reason we have exposed. On the other we see if this virtual adversarial acts as a good regularizer and thus the objective is see if it can substitute the gaussian noise by adversarial. For the labeled data we compute UA and add it in each layer. \\

To analyze this we have done the next experiments. On one hand to check if we can help minimizing the MSE we add UA only to the input of the unlabeled data. On the other we add multilayer adversarial. We first present the experiments checking if the adversarial can be used as a regularizer. For this we check to compute the norm and the sign of the gradient in several ways:

\begin{itemize}
    \item Fix factor
    \item Drop Factor
    \item Drop Factor plus gaussian noise
\end{itemize}

Fix factor implies setting a fix value for $\tau$. We check for the next values: 0.05,0.1,0.15,0.3. The drop factor implies dropping the factor in each iteration by defining a normal distribution with standard deviation given by: 0.05,0.1,0.15,0.3. The reasons for doing this dropping is the next one. Approximatetly the 70\% of values of a normal distribution lies between $[-\sigma,\sigma]$. This means that the norm of a gaussian noise vector is upper bounded by $\sqrt{\underset{k}{\sum}\sigma^2}$, where $k$ is the dimension of the vector, in the 60\% of the dropped values (remark that this percentage is in case all the dimensions in the range are exactly $\lvert\sigma\rvert$, if they are lower the percentage increases.)\todo{Esto se justifica bien con una binomial, por si os suena a cuento de la lechera}. If we drop the $\tau$ according to this we will created adversarial vectors similar in power to the ones we drop when generating gaussian noise. Having a fix value for this factor would no be a good candidate for substituting gaussian noise. Moreover the norm adversarial is more similar to the gaussian vector as the sign adversarial have a fix value for the elements of the vector.\\

When performing our experiments we notice several things. The first one is that the ladder network is better regularized when using droped factor vs the fix factor, as we expected. Moreover what better regularize is mixing gaussian noise and UA. Note that the adversarial noise push the features towards decision bounds in which the cost is increased and this is the closest inbetween class bounds. However we can have bounds respect to other classes in other directions and pushing towards that direction can also help the ladder network extract relevant features to better generalize. On the other hand the experiments without gaussian noise where better with the sign. This is because when truncating the phase of the adversarial vector we give some randomness to this vector and we know that the ladder network is very sensible to gaussian noise to perform good (both supervised and unsupervised learning). This also matches with what we saw when mixing UA with Gaussian. However the norm adversarial goes better when we incorporate the gaussian noise. We also show the results of adding UA only to unlabeled samples and same to labeled samples to check how this new way of regularize perform in different targets (Cross entropy vs MSE). We can get state of art results as shown in the next tables.\todo{Comentar el 50 etiquetas que habia mucho outlier}:

\begin{table}[H]
\centering
\begin{adjustbox}{max width=\textwidth}
\begin{tabular}{|l|l|l|l|l|}
\hline
\multicolumn{5}{|c|}{\textbf{FC MNIST}}                                                        \\ \hline
\textbf{\# of labels} &                 50       & 100  & 1000  & All labels                           \\ \hline
Ladder full  \citep{DBLP:journals/corr/RasmusVHBR15}    &    1.750\% $\pm$ 0.582     & 1.111\% $\pm$ 0.105 & 0.880\% $\pm$ 0.075& 0.654\% $\pm$ 0.052 \\ \hline
\textbf{Adversarial Ladder full} & 1.417\% $\pm$ 0.539  & 1.110\% $\pm$ 0.055 &  0.861\% $\pm$ 0.071 & 0.639\% $\pm$ 0.056 \\ \hline
\textbf{Unsupervised Adversarial Ladder full} & 1.295\% $\pm$ 0.391  & 1.096\% $\pm$ 0.058 &  0.92\% $\pm$ 0.086 & 0.637\% $\pm$ 0.050 \\ \hline
\textbf{UA unlabeled Ladder full} & 1.439\% $\pm$ 0.5652  & 1.126\% $\pm$ 0.058 &  0.922\% $\pm$ 0.072 & 0.614\% $\pm$ 0.028 \\ \hline
\textbf{UA labeled Ladder full} & 1.849\% $\pm$ 0.645  & 1.131\% $\pm$ 0.071 &  0.908\% $\pm$ 0.082 & 0.645\% $\pm$ 0.044\\ \hline
\end{tabular}
\end{adjustbox}
\caption{Results for fully connected MNIST unsupervised Adversarial. UA labeled refers to adding unsupervised adversarial only to labeled samples}
\label{tab:MNISTFC}
\end{table}

\begin{table}[H]
\centering
\begin{adjustbox}{max width=\textwidth}
\begin{tabular}{|l|l|l|l|l|}
\hline
\multicolumn{5}{|c|}{\textbf{FC MNIST}}                                                        \\ \hline
\textbf{Unsupervised Adversarial Ladder full} & (n;$0.1$)  & (n;$0.07$) & (n;$0.19$)  & (n;$0.3$) \\ \hline
\textbf{UA unlabeled Ladder full} & (n;$0.1$)  & (n;$0.05$) & (n;$0.05$)  & (n;$0.05$) \\ \hline
\textbf{UA labeled Ladder full} & (n;$0.05$) & (n;$0.3$) & (n;$0.05$) & (n;$0.1$)\\ \hline
\end{tabular}
\end{adjustbox}
\caption{Configurations of the adversarial noise addition of the above table}
\label{tab:MNISTFC}
\end{table}

As we can see in the above table, the results for the 50, 100 and 60000 tags  are better than the supervised adversarial framework. Moreover the 100 tag and 1000 tag needs more refinement as we had to do in supervised adversarial.

\begin{table}[H]
\centering
\begin{adjustbox}{max width=\textwidth}
\begin{tabular}{|l|l|l|l|}
\hline
\multicolumn{2}{|c|}{\textbf{Convolutional MNIST}}                                           \\ \hline
\textbf{\# of labels}                 & 100 (n;$10^{-3}$)                                  \\ \hline
Conv-Small, $\Gamma$-model \citep{DBLP:journals/corr/RasmusVHBR15}          & 0.938\%  $\pm$ 0.423                                \\ \hline
\textbf{Adversarial Ladder $\Gamma$-model}          &  0.884\% $\pm$ 0.427                                         \\ \hline
\textbf{Unsupervised Adversarial Ladder $\Gamma$-model}          &  0.86\% $\pm$ 0                                          \\ \hline
\end{tabular}
\end{adjustbox}
\caption{Results for Convolutional MNIST unsupervised Adversarial}
\label{tab:MNISTCONV}
\end{table}

The result reported for Unsupervised Adversarial Ladder convolutional model correspond only to the result on one seed, as the models for the other one suffer from numeric saturation. This seed model in the supervised adversarial had a 0.76\% accuracy. We end up this subsection showing the results of adding only unsupervised adversarial to the input of the unlabeled part of the network. We explore if the unsupervised adversarial can help in minimizing the MSE, as the hidden regularization is done with gaussian noise. The next table show the results. For this experiment we did not perform refinement:

\begin{table}[H]
\centering
\begin{adjustbox}{max width=\textwidth}
\begin{tabular}{|l|l|l|l|l|}
\hline
\multicolumn{5}{|c|}{\textbf{FC MNIST}}                                                        \\ \hline
\textbf{\# of labels} &                 50       & 100  & 1000  & All labels                           \\ \hline
Ladder full  \citep{DBLP:journals/corr/RasmusVHBR15}    &    1.750\% $\pm$ 0.582     & 1.111\% $\pm$ 0.105 & 0.880\% $\pm$ 0.075& 0.654\% $\pm$ 0.052 \\ \hline
\textbf{Adversarial Ladder full} & 1.417\% $\pm$ 0.539  & 1.110\% $\pm$ 0.055 &  0.861\% $\pm$ 0.071 & 0.639\% $\pm$ 0.056 \\ \hline
\textbf{UA input unlabeled Ladder full DF} & 1.678\% $\pm$ 0.555  & 1.152\% $\pm$ 0.069 &  0.914\% $\pm$ 0.071 & 0.606\% $\pm$
0.043 \\ \hline
\textbf{UA input unlabeled Ladder full FF} & 1.397\% $\pm$ 0.485  & 1.114\% $\pm$ 0.078 &  0.885\% $\pm$ 0.087 & 0.625\% $\pm$ 0.042 \\ \hline
\end{tabular}
\end{adjustbox}
\caption{Results for fully connected MNIST unsupervised Adversarial only added to the input of the unlabeled network part. FF means fix factor and DF drop factor.}
\label{tab:MNISTFC}
\end{table}

\end{subsection}

\end{section}
\begin{section}{Conclusions and Future Work}

We have shown results on adding adversarial noise to the ladder network, in addition of new possibilities for semisupervised task. One of the key things adversarial helps for is in overfitted model. Moreover we see how unsupervised adversarial can help semi supervised task such as a regularizator with the gaussian noise and performing good in minimizing the MSE. For the first case the drop factor framework went better and for the second was the fix factor. Dropping factor used for regularization make more sense as the generated UA noise is similar in power distribution to the gaussian noise. On the other hand when pushing samples to try and denoise them seams better to have a fix value to take out this variation and let the network learn to denoise (remember this perturbation is in a really sensible direction). \\

Improving the results on the 50 tag was easy, we did not perform fine tunning of the hyperparameters. This was not the case with the 100 tag. On the other hand the experiments with unsupervised adversarial did not require hyperparameter search (except some cases), as we can see in the tables. Our first hyperparameter where chosen taking in account the standard deviation of the gaussian noise.\\

Future work will be focused in finishing the experiment of supervised multilayer adversarial and mixing the different proposals of this work.

\end{section}

\bibliographystyle{abbrvnat}
%\bibliography{main_definitivo}

\end{document}